\documentclass{article}

\usepackage{microtype}
\usepackage{graphicx}
\usepackage{subcaption}
\usepackage{booktabs} 

\usepackage{hyperref}

\usepackage[accepted]{icml2026}

\usepackage{amsmath}
\usepackage{amssymb}
\usepackage{mathtools}
\usepackage{amsthm}

\usepackage{multirow}    
\usepackage{array}       

\newcommand{\cmark}{\checkmark}  
\newcommand{\xmark}{\ding{55}}   
\usepackage{pifont}             

\newcommand{\tblidx}[1]{\textbf{#1}}

\usepackage[table]{xcolor} 

\definecolor{lightgray}{gray}{0.8}

\usepackage[capitalize,noabbrev]{cleveref}

\theoremstyle{plain}

\theoremstyle{definition}

\theoremstyle{remark}

\usepackage[textsize=tiny]{todonotes}

\icmltitlerunning{SMM Transformer: Leveraging SNNs for Multimodal Tasks}

\begin{document}

\twocolumn[
  \icmltitle{SMM Transformer: Leveraging Spiking Neural Networks for Multimodal Tasks}



  \icmlsetsymbol{equal}{*}

  \begin{icmlauthorlist}
    \icmlauthor{Xiubo Liang}{yyy}
    \icmlauthor{Jinxing Han}{yyy}
    \icmlauthor{Yuke Li}{comp}
    \icmlauthor{Haoqi Zhu}{comp}
    \icmlauthor{Yu Zhao}{yyy}
    \icmlauthor{Hongzhi Wang}{yyy,comp}
  \end{icmlauthorlist}
  \icmlaffiliation{yyy}{School of Software Technology, Zhejiang University, Ningbo, China}
  \icmlaffiliation{comp}{NetEase Yidun AI Lab, Hangzhou, China}

  \icmlcorrespondingauthor{Hongzhi Wang}{hongzhiwang@zju.edu.cn}
  
  \icmlkeywords{Machine Learning, ICML}

  \vskip 0.3in
]



\printAffiliationsAndNotice{}  

\begin{abstract}
    Spiking Neural Networks (SNNs) enable event-driven computation with sparse activations, but building multimodal Transformers on SNNs is hindered by unstable training in deep spiking stacks and the mismatch between dense softmax attention and spike-based communication. We propose SMM Transformer, an SNN-based multimodal Transformer framework that combines (i)PLMP, a Parallel LIF with Multistage Learnable Parameters neuron and a tailored P-STBP algorithm for stable deep SNN training, (ii) SMSA, an attention-inspired spike-driven token-mixing module that replaces dense pairwise softmax attention with channel-wise spike co-activation and self-compensation, and (iii)SMoE, a spiking mixture-of-experts module for modality-aware fusion. Across visual and multimodal benchmarks, SMM Transformer achieves competitive accuracy compared to ANN baselines. Under a standard MAC/AC arithmetic model, SMSA reduces the estimated operator-level compute energy of the attention module by up to 97\%, while whole-model profiling shows more moderate but consistent efficiency gains.
\end{abstract}

\section{Introduction}
Large language models (LLMs) and multimodal foundation models have rapidly advanced vision--language learning, yet their training and deployment costs motivate research into more energy-efficient alternatives \cite{LLM1,LLM2,LLM3,grattafiori2024llama3herdmodels}.
In this work, we do not aim to scale to LLM-sized architectures; instead, we target the dominant compute bottleneck in Transformer-style multimodal systems---dense attention---and study how event-driven spiking computation can reduce its operator-level energy cost.
By contrast, spiking neural networks (SNNs) emulate neuronal firing with sparse binary spikes and offer a promising substrate for energy-efficient computing research \cite{2watts}.\par

Multimodal learning underpins assistive vision and large-scale media annotation. Modern vision--language models typically rely on dense Transformer attention to align image and text representations, which is accurate but compute intensive \cite{sa1,sa2,sa3,sa4,sa5,sa6}.
On the SNN side, prior work has largely focused on single-modality perception and spiking backbones \cite{spikformer}, while multimodal vision--language SNNs remain underexplored and often rely on separate encoders or late fusion that limits early cross-modal interaction \cite{wang2025cross,nnc1,nnc2,nnc3}.
These gaps motivate a Transformer-inspired spiking architecture that supports efficient fusion under sparse spiking computation.\par

\begin{figure*}[!t]
    \centering
    \includegraphics[width=0.975\linewidth, keepaspectratio]{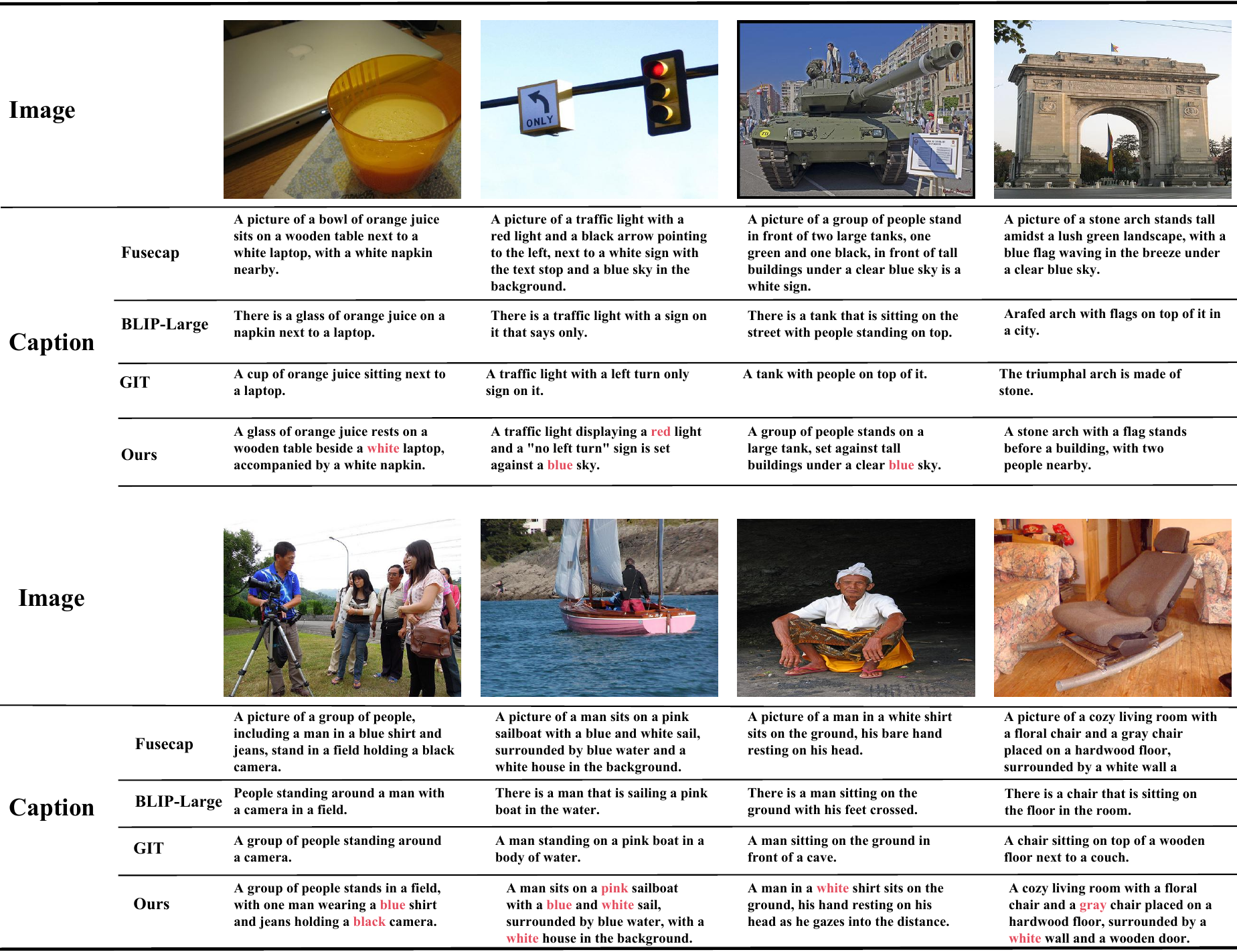}
    \caption{Comparison of image captioning results from different models in the image shows that our SMM Transformer provides more accurate descriptions. Additionally, it captures color features well.}
    \label{fig:results}
\end{figure*}

This study explores Transformer--SNN hybrids for high-accuracy and energy-aware multimodal learning. To mitigate gradient attenuation and tedious parameter tuning in LIF-based SNNs \cite{LifNode}, we introduce the PLMP neuron. PLMP uses multiple learnable LIF branches to capture heterogeneous temporal dynamics, followed by a fixed-threshold read-out gate that produces a binary spike. 
We further tailor Spatio-Temporal Back-propagation (STBP) \cite{STBP} into P-STBP to propagate gradients through the multi-branch temporal dynamics and the read-out gate. \par

For token mixing, we propose SMSA, a PLMP-based spike-driven module inspired by self-attention. SMSA replaces the dense $N\times N$ token-pair matrix with spike co-activation, sequence-level channel aggregation, value masking, and a lightweight self-compensation branch. This design trades part of the dense token-to-token inductive bias for sparse, event-driven computation. For multimodal fusion, we introduce a Spiking Mixture-of-Experts (SMoE), which allocates modality-specific feed-forward capacity to vision, language, and vision--language experts while maintaining a shared interaction pathway. \par

We retain the Transformer macro-architecture with stacked residual blocks, but redesign the spiking neuron, token-mixing, and feed-forward fusion components to be compatible with sparse spike communication. Experimental results demonstrate that SMM Transformer achieves accuracy comparable to traditional ANNs in both visual tasks and multimodal tasks. We also designed ablation experiments to validate the contribution of each module to the overall performance. Our main contributions are summarized as follows:
\begin{itemize}
    \item \textbf{Trainable spiking unit.} We propose PLMP, a parallel multistage LIF unit with learnable membrane time constants and thresholds, and introduce P-STBP to stably train PLMP-based deep spiking stacks.
    \item \textbf{Spike-driven token mixing.} We design SMSA, an attention-inspired token-mixing module built on PLMP read-out spikes. SMSA replaces dense pairwise softmax attention with channel-wise spike co-activation, sparse masking, and self-compensation.
    \item \textbf{Modality-aware fusion.} We introduce SMoE, which routes tokens to vision, language, or vision--language experts to reduce cross-modal interference while keeping a shared interaction backbone.
    \item \textbf{Evaluation and analysis.} We validate SMM on visual and vision--language benchmarks with ablations, and provide an operator-level energy analysis of attention under the standard MAC/AC model.
\end{itemize}

\section{Related Works}

\begin{figure*}[!t]
    \centering
    \includegraphics[width=0.99\linewidth]{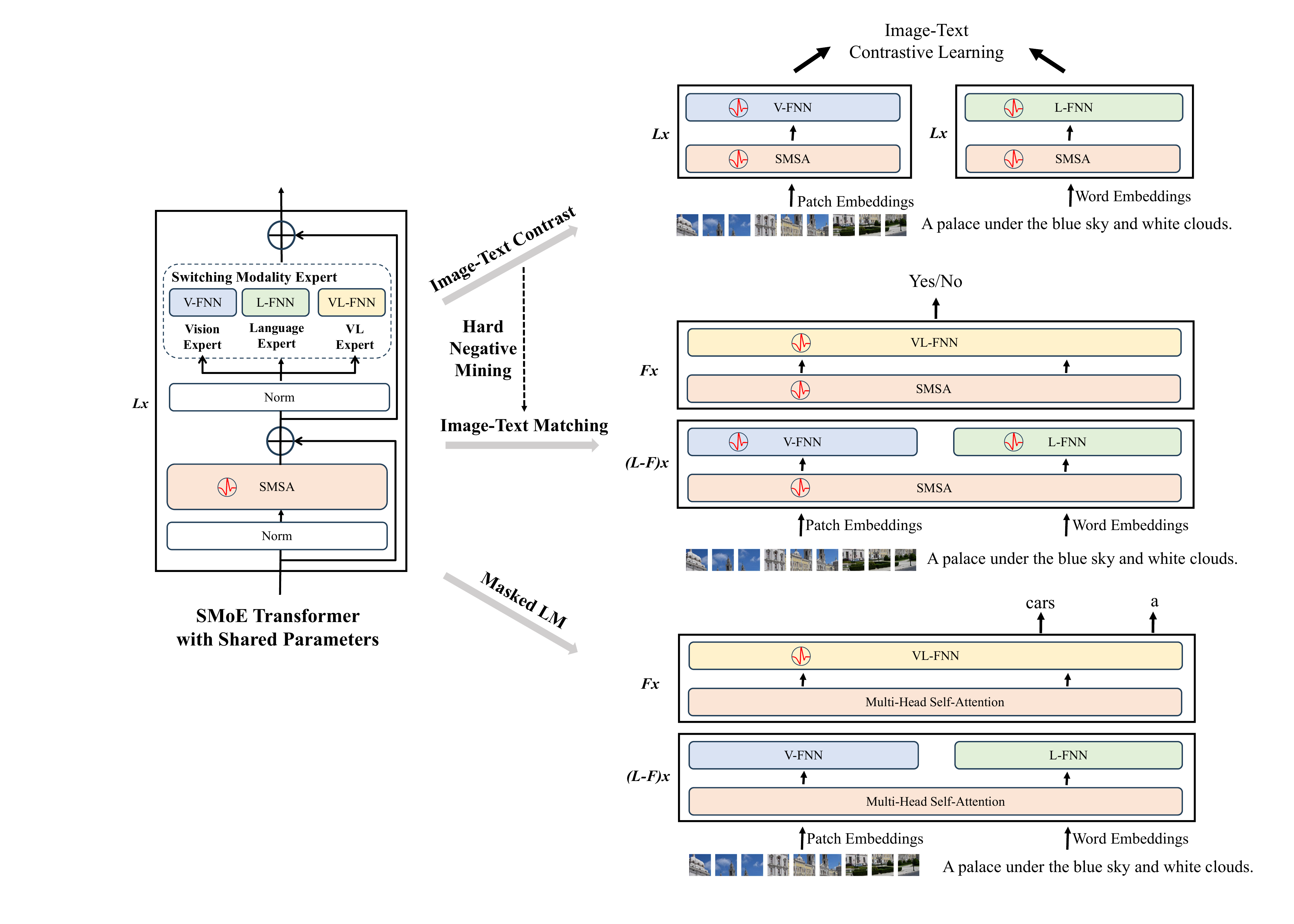}
    \caption{Overview of SMM Transformer. The model parameters are shared across image-text contrastive learning, masked language modeling, and image-text matching pre-training tasks. }
    \label{fig:SMOE}
\end{figure*}

\subsection{SNN Training Methods}
Training strategies for SNNs are broadly classified into unsupervised and supervised paradigms.  
Unsupervised learning emulates biological synaptic plasticity, exemplified by Hebbian learning \cite{hebb} and spike-timing-dependent plasticity (STDP) \cite{STDP}.  
Supervised techniques comprise (i) indirect ANN-to-SNN conversion \cite{ANN2SNN1,ANN2SNN3}, which often sacrifices accuracy, and (ii) direct gradient-based algorithms that circumvent spike non-differentiability, such as SpikeProp \cite{spikeprop} and STBP \cite{STBP}.  
Subsequent advances extend network depth (e.g., SEW-ResNet \cite{SEW-Resnet}) and integrate spiking blocks within Transformer frameworks for vision tasks \cite{spikformer,Spike-driven}, yielding performance comparable to state-of-the-art CNNs \cite{DSNN3,DSNN7,DSNN4,DSNN5}.  
Consequently, direct training has become the prevailing approach, promoting efficient end-to-end optimization.

\subsection{Multimodal Tasks}

\subsubsection{Image Caption}
Image captioning has evolved from handcrafted descriptors—SIFT \cite{35} and HOG \cite{36}—to CNN–RNN architectures endowed with spatial attention, exemplified by Show, Attend and Tell \cite{37} and Bottom-Up and Top-Down Attention \cite{38}.  
Transformer backbones \cite{Transformer} further strengthen relational reasoning—Object Relation Transformer \cite{39} and X-Linear attention \cite{40}—and enable fully detector-free, end-to-end pipelines \cite{41}.  
Complementary paradigms, including GAN-based training \cite{gan}, reinforcement learning \cite{rl}, and dense captioning \cite{44}, alleviate exposure bias while enriching object semantics.

\subsubsection{Retrieval}
Cross-modal retrieval seeks a shared embedding space for visual and textual modalities.  
In image-to-text retrieval, contrastive objectives align paired samples; attention or graph representations refine object-level correspondence \cite{xie2024image}, whereas attribute-centric descriptions bolster zero-shot generalisation \cite{zeng2024meacap}.  
Conversely, text-to-image retrieval employs dual-stream encoders \cite{suo2024knowledge} and diffusion priors \cite{koley2024text} to capture nuanced semantics. Techniques such as knowledge injection \cite{suo2024knowledge}, implicit concept mining \cite{wang2025cross}, and multi-head hashing \cite{liu2024multi} further fortify robustness and scalability.

\renewcommand{\dblfloatpagefraction}{.9}
\begin{figure*}
    \centering
    \includegraphics[width=\linewidth]{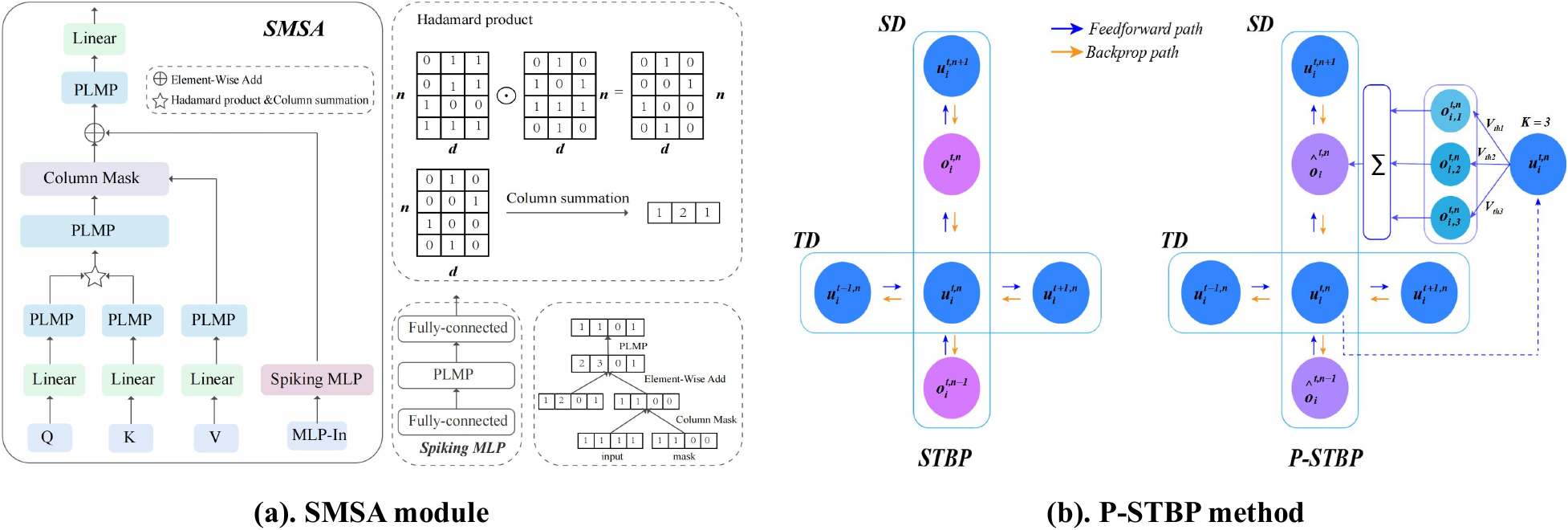}
    \caption{Illustration of the proposed components. (a) SMSA first maps Query, Key, and Value into PLMP read-out spikes, then performs spike co-activation, column-wise aggregation, value masking, and self-compensation without constructing a dense $N\times N$ attention matrix. (b) P-STBP propagates gradients through both the temporal branch dynamics and the fixed-threshold PLMP read-out gate.}
    \label{fig:components}
\end{figure*}

\section{Method}
As summarized in Fig.~\ref{fig:SMOE} and Fig.~\ref{fig:components}, SMM follows a Transformer-style residual stack, but replaces key building blocks to fit spiking computation. PLMP serves as the trainable spiking unit throughout the network, SMSA provides attention-inspired spike-driven token mixing without forming a dense $N\times N$ attention matrix, and SMoE injects modality-aware capacity in the feed-forward path while sharing a unified fusion backbone. We next describe each component and then detail how they are trained and evaluated in the multimodal pipeline. \par

\subsection{PLMP}
In conventional SNN training with LIF neurons, the membrane time constant ($\tau$) and voltage threshold ($V_{th}$) are treated as hyper-parameters that must be hand-tuned or optimised. Such manual specification is biologically implausible: $\tau$—the time for the membrane potential $u$ to equilibrate—should scale with membrane resistance ($R_{m}$) and capacitance ($C_{m}$). Additionally, the $V_{th}$, influenced by spiking input and discharge history, exhibits high variability both within and between cells. Computationally, forward propagation reduces to accumulating weighted presynaptic spikes and comparing the resulting membrane voltage with $V_{th}$. Consequently, empirical hyper-parameter selection inflates trial-and-error cost.

To overcome these limitations, we propose the PLMP neuron, whose temporal leakage and firing thresholds are learnable. Each PLMP unit contains $K$ parallel LIF branches. The $k$-th branch has its own membrane state $u_{i,k}^{t,n}$, learnable leakage factor $\alpha_k$, and learnable threshold $V_{th,k}$. Given the input current $r_i^{t,n}$, the branch dynamics are
\begin{equation}
    u_{i,k}^{t,n} = \alpha_k u_{i,k}^{t-1,n}(1-o_{i,k}^{t-1,n}) + r_i^{t,n},
\end{equation}
\begin{equation}
    o_{i,k}^{t,n} = \mathcal{H}(u_{i,k}^{t,n}-V_{th,k}),
\end{equation}
where $\mathcal{H}(\cdot)$ is the Heaviside spike function and $o_{i,k}^{t,n}\in\{0,1\}$ is the branch spike. The $K$ branch spikes are then aggregated into an internal spike count
\begin{equation}
    c_i^{t,n} = \sum_{k=1}^{K} o_{i,k}^{t,n}.
\end{equation}
PLMP propagates only a binary read-out spike to the next layer:
\begin{equation}
    \hat{o}_i^{t,n} = \mathcal{H}(c_i^{t,n}-\theta_{\mathrm{ro}}), \quad \theta_{\mathrm{ro}}=K.
\end{equation}
Thus, PLMP is not a stack of two unrelated LIF neurons. The parallel branches model heterogeneous temporal responses, while the fixed-threshold read-out gate performs coincidence-style binarization. Unless otherwise stated, the output of PLMP denotes the binary read-out spike $\hat{o}_i^{t,n}$ rather than the internal spike count $c_i^{t,n}$.

\subsection{P-STBP}
Given that PLMP units have different computational dynamics from standard LIF neurons, their forward propagation and back-propagation also differ. We therefore use P-STBP, a PLMP-specific extension of STBP tailored to multi-branch learnable membrane dynamics. Conceptually, P-STBP jointly propagates errors along spatial and temporal dimensions. As shown in Fig.~\ref{fig:components}(b), the spatial path transfers gradients across layers through PLMP read-out spikes, while the temporal path propagates gradients through the recurrent membrane states of parallel LIF branches.

For a connection from neuron $j$ in layer $n-1$ to neuron $i$ in layer $n$, the weight gradient accumulates the contributions from all simulation steps and all PLMP branches:
\begin{align}
    \frac{\partial L}{\partial w_{ij}^{n}}
    &=
    \sum_{t=1}^{T}
    \left(
    \sum_{k=1}^{K}
    \frac{\partial L}{\partial u_{i,k}^{t,n}}
    \frac{\partial u_{i,k}^{t,n}}{\partial r_i^{t,n}}
    \right)
    \frac{\partial r_i^{t,n}}{\partial w_{ij}^{n}}
    \notag \\
    &=
    \sum_{t=1}^{T}
    \sum_{k=1}^{K}
    \frac{\partial L}{\partial u_{i,k}^{t,n}}
    \hat{o}_{j}^{t,n-1},
\end{align}
where $r_i^{t,n}$ is the input current of neuron $i$ at time step $t$, $w_{ij}^{n}$ is the synaptic weight, and $\hat{o}_{j}^{t,n-1}$ is the binary read-out spike from the previous layer. Similarly, the bias gradient is
\begin{equation}
    \frac{\partial L}{\partial b_i^{n}}
    =
    \sum_{t=1}^{T}
    \sum_{k=1}^{K}
    \frac{\partial L}{\partial u_{i,k}^{t,n}}.
\end{equation}

The learnable leakage factor controls the contribution of the previous membrane state to the next membrane state. Its gradient is computed as
\begin{equation}
    \frac{\partial L}{\partial \alpha_k}
    =
    \sum_{n,i,t}
    \frac{\partial L}{\partial u_{i,k}^{t+1,n}}
    u_{i,k}^{t,n}
    \left(
    1-o_{i,k}^{t,n}
    \right),
\end{equation}
where $\alpha_k$ is the learnable leakage factor of the $k$-th branch and $o_{i,k}^{t,n}$ is the corresponding branch spike.

The learnable threshold is updated through the surrogate derivative of the branch spike function:
\begin{equation}
    \frac{\partial L}{\partial V_{th,k}}
    =
    -
    \sum_{n,i,t}
    \frac{\partial L}{\partial o_{i,k}^{t,n}}
    \phi_{\mathrm{br}}
    \left(
    u_{i,k}^{t,n}
    -
    V_{th,k}
    \right),
\end{equation}
where $V_{th,k}$ is the threshold of the $k$-th branch and $\phi_{\mathrm{br}}(\cdot)$ denotes the surrogate derivative used for the branch spike function.

These gradients show the core difference between P-STBP and standard STBP. Standard STBP propagates through a single membrane trajectory, whereas P-STBP aggregates gradients from $K$ parallel LIF branches and jointly optimizes synaptic weights, biases, leakage factors, and firing thresholds. Additional error-propagation details are provided in Appendix~\ref{app:plmp_pstbp}.
\subsection{SMSA}
Building upon PLMP, we propose Spiking MLP Self-Attention (SMSA), an attention-inspired spike-driven token-mixing module that avoids constructing the dense $N\times N$ attention matrix. SMSA does not exactly reproduce softmax self-attention. Instead, it replaces pairwise token affinity with sequence-level channel-wise spike co-activation, which provides a practical accuracy-efficiency trade-off for SNN-based multimodal learning.

As illustrated in Fig.~\ref{fig:components}(a), we first project the input $X\in\mathbb{R}^{N\times D}$ and apply PLMP read-out neurons to obtain binary spike tensors:
\begin{align}
    Q_s &= \operatorname{PLMP}(XW^Q), \notag \\
    K_s &= \operatorname{PLMP}(XW^K), \notag \\
    V_s &= \operatorname{PLMP}(XW^V).
\end{align}
where $Q_s,K_s,V_s\in\{0,1\}^{N\times D}$. Instead of computing $QK^\top$ followed by softmax, SMSA first estimates channel-wise spike co-activation:
\begin{equation}
    G_c = SUM_c(Q_s \odot K_s),
\end{equation}
where $SUM_c(\cdot)$ aggregates over the token dimension and produces a $1\times D$ channel vector. We then binarize this channel vector with PLMP to obtain a spike gate:
\begin{equation}
    G_s = \operatorname{PLMP}(G_c).
\end{equation}
The gate is applied to the value spikes through column-wise masking:
\begin{equation}
    M_s = G_s \otimes V_s.
\end{equation}
To compensate for information loss caused by spike binarization and the removal of dense token-pair interactions, we add a parallel self-compensation branch $SM(X)$ implemented as a spiking MLP:
\begin{equation}
    Z_s = \operatorname{PLMP}(M_s \oplus SM(X)).
\end{equation}
Finally, the SMSA output is obtained by a linear output projection:
\begin{equation}
    SMSA(X) = Z_sW^O.
\end{equation}
Here, $\odot$ denotes the Hadamard product, $\otimes$ denotes column-wise masking, and $\oplus$ denotes element-wise summation. The sparse mixing path operates on binary PLMP read-out spikes, while the final projection consumes the binary tensor $Z_s$.

Compared with vanilla self-attention, SMSA removes the $O(N^2)$ token-pair attention matrix and replaces it with channel-wise spike co-activation and masking. This design reduces arithmetic cost and enables sparse accumulate-based computation when spike activations are sparse. The trade-off is that SMSA no longer preserves the exact token-to-token inductive bias of dense attention. The self-compensation branch is therefore used to recover part of the lost expressiveness, which is empirically validated in Sec.~\ref{sec:ablation}.

\begin{figure}
    \centering
    \includegraphics[width=\linewidth]{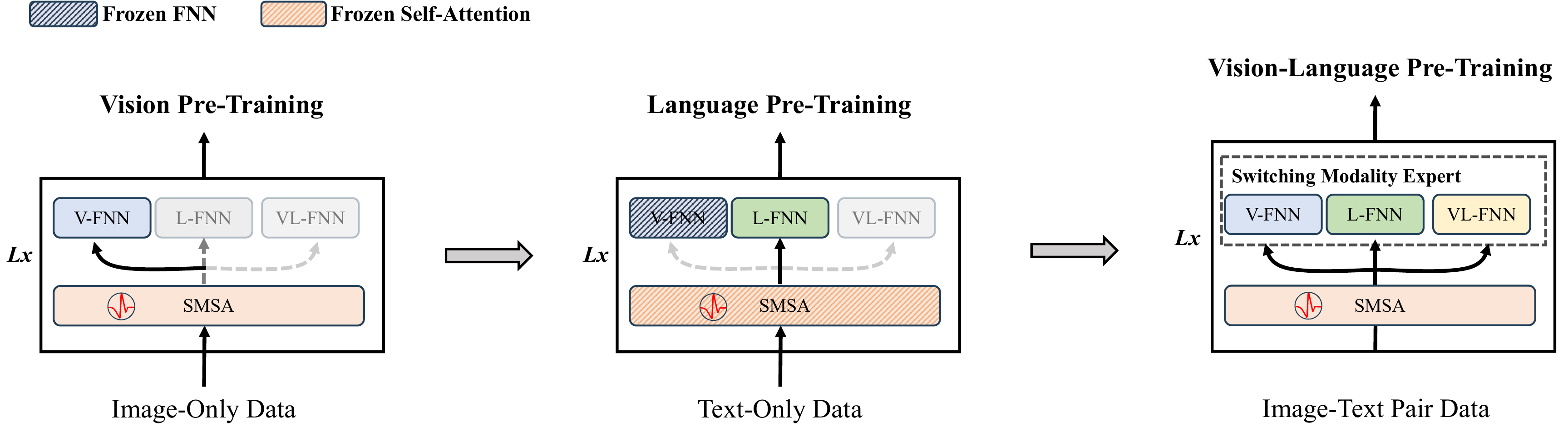}
    \caption{Illustration of stagewise pre-training strategy.}
    \label{fig:frozen}
\end{figure}
\begin{equation}
    L_R(\theta) = - E_{y_{1:t} \sim p_\theta [r(y_{1:t})]},
\end{equation}
$r(y_{1:t})$ is usually non-differentiable, the gradient of the  $L_R(\theta)$ can be described by Eq.\ref{lr}.
\begin{equation}
\label{lr}
    \nabla L_R(\theta) \approx (r(y_s^{1:T}) - r(\hat{y}_{1:T})) \nabla \log p_\theta (y_s^{1:T}),
\end{equation}
where $y_s^{1:T}$ is a descriptive sample. $r(y_s^{1:T})$ is the greedy decoding score obtained from the current model. \par

\subsection{SMoE}
In this section, we introduce SMoE, a spike-driven mixture-of-experts framework that dynamically selects modality-specific experts. SMoE is designed to preserve the representation benefit of expert routing while keeping the fusion pathway compatible with sparse spiking computation. Its structure is
\begin{equation}
    y = \sum_{k=1}^{K} R_k(X_A) \cdot E_k(X_A),
\end{equation}
where $X_A$ is the input sequence, $R_k(X_A)$ is the routing weight for the $k$-th expert, and $E_k(X_A)$ is the output of the corresponding spiking expert.

The routing mechanism selects the top-$K$ experts as
\begin{equation}
    R(X_A) = \operatorname{TopK}(\operatorname{softmax}(X_AW_R)).
\end{equation}
Because the routing is driven by spike-based representations, expert utilization is sparse and modality-aware. As shown in Fig.~\ref{fig:SMOE} and Fig.~\ref{fig:frozen}, we use a stagewise pre-training strategy. We first pretrain the vision expert and the shared SMSA module on image-only data. We then freeze these components and train the language expert with masked language modeling on text-only data. Finally, we perform vision-language pre-training to jointly optimize the full model on image-text pairs.

\subsection{Image-Text Representations}
\subsubsection{Image Representations}
Following vision Transformers \cite{vit}, the 2D image \( \mathbf{v} \in \mathbb{R}^{H \times W \times C} \) is split and reshaped into \( N = \frac{HW}{P^2} \) patches, where \( (H, W) \) is the resolution of the input image, and \( P \) is the patch resolution. The image patches are then flattened into vectors and linearly projected to obtain patch embeddings. We prepend a learnable special token \([T_{\text{CLS}}]\) to the sequence. Finally, the image input representations are obtained by summing the patch embeddings, learnable 1D position embeddings \( V_{\text{pos}} \in \mathbb{R}^{(N+1) \times D} \), and image type embeddings \( V_{\text{type}} \in \mathbb{R}^D \):
\begin{equation}
H_0^{\mathbf{v}} = [v_{[T_{\text{CLS}}]}, v_1^p, \dots, v_N^p] + V_{\text{pos}} + V_{\text{type}},
\end{equation}
where \( H_0^{\mathbf{v}} \in \mathbb{R}^{(N+1) \times D} \) and the linear projection \( V \in \mathbb{R}^{(P^2 C) \times D} \).

\subsubsection{Text Representations}
Following BERT \cite{bert}, we tokenize the text to subword units using WordPiece. A start-of-sequence token \([T_{\text{CLS}}]\) and a special boundary token \([T_{\text{SEP}}]\) are added to the text sequence. Text input representations \( H_0^{\mathbf{w}} \in \mathbb{R}^{(M+2) \times D} \) are computed by summing the corresponding word embeddings, text position embeddings, and text type embeddings:
\begin{equation}
    H_0^{\mathbf{w}} = [w_{[T_{\text{CLS}}]}, w_1, \dots, w_M, w_{[T_{\text{SEP}}]}] + T_{\text{pos}} + T_{\text{type}},
\end{equation}
where \( M \) indicates the length of tokenized subword units.

\subsubsection{Image-Text Representations}
Image and text input vectors are concatenated to form the image-text representation:
\begin{equation}
    H_0^{\mathbf{vl}} = [H_0^{\mathbf{w}}, H_0^{\mathbf{v}}],
\end{equation}
where \( H_0^{\mathbf{vl}} \) denotes the joint image-text representation.

\section{Experiments}

\subsection{Comparative Quantitative Experiments}
\subsubsection{Retrieval}

\begin{table*}[!t]
\centering
\caption{Fine-tuning results of text-retrieval (TR) and image-retrieval (IR) on COCO and Flickr30K. ALBEF first encodes images and text separately to obtain the top-k candidates, and then feed these representations into a fusion encoder to rerank the candidates. The others require to encode all image-text combinations by a fusion encoder.}
\label{retrieve}
\resizebox{0.95\textwidth}{!}{
    \begin{tabular}{l|cccccc|cccccc}
    \hline
    \textbf{Model} & \multicolumn{6}{c|}{\textbf{MSCOCO (5K test set)}} & \multicolumn{6}{c}{\textbf{Flickr30K (1K test set)}} \\
    \cline{2-13}
     & R@1 & R@5 & R@10 & R@1 & R@5 & R@10 & R@1 & R@5 & R@10 & R@1 & R@5 & R@10 \\
    \hline
    \textbf{Base-Size Models} & & & & & & & & & & & \\
    UNITER-Base & 64.4 & 87.4 & 93.1 & 50.3 & 78.5 & 87.2 & 85.9 & 97.1 & 98.8 & 72.5 & 92.4 & 96.1 \\
    VILLA-Base & - & - & - & - & - & - & 86.6 & 97.9 & 99.2 & 74.7 & 92.9 & 95.8 \\
    ViLT-Base & 61.5 & 86.3 & 92.7 & 42.7 & 72.9 & 83.1 & 83.5 & 96.7 & 98.6 & 64.4 & 88.7 & 93.8 \\
    ALBEF-Base & 73.1 & 91.4 & 96.0 & 56.8 & 81.5 & 89.2 & 94.3 & 99.4 & 99.8 & 82.8 & 96.7 & 98.4 \\
    SMM-Transformer-Base & 72.8 & 91.4 & 95.5 & 56.4 & 81.3 & 88.8 & 91.5 & 99.1 & 99.9 & 78.4 & 94.5 & 96.1 \\
    \hline
    \textbf{Large-Size Models} & & & & & & & & & & & \\
    UNITER-Large & 65.7 & 88.6 & 93.8 & 52.9 & 79.9 & 88.0 & 87.3 & 98.0 & 99.2 & 75.6 & 94.1 & 96.8 \\
    VILLA-Large & - & - & - & - & - & - & 87.9 & 97.5 & 98.8 & 76.3 & 94.2 & 96.8 \\
    SMM-Transformer-Large & \textbf{75.3} & \textbf{93.1} & \textbf{96.1} &\textbf{ 59.3} & \textbf{82.1} & \textbf{90.2} & \textbf{93.9} &\textbf{ 99.9 }& \textbf{100.0} &\textbf{ 83.7} &\textbf{ 95.8} & \textbf{97.2 }\\
    \hline
    \end{tabular}
}
\end{table*}

\begin{table*}[t]
    \centering
    \caption{Image captioning results on MSCOCO (Karpathy split). We report standard captioning metrics and measured power.}
    \scalebox{0.85}{
    \begin{tabular}{@{}lcccccccccc@{}}
    \toprule
    Model & Type & BLEU-1 & BLEU-2 & BLEU-3 & BLEU-4 & METEOR & ROUGE-L & CIDEr & SPICE & Energy (mJ) \\ \midrule
    SCST & ANN & 78.1 & 61.9 & 47.0 & 35.2 & 27.0 & 56.3 & 114.7 & - & 102.3 \\
    RFNet & ANN & 79.1 & - & - & 36.5 & 27.7 & 57.3 & 121.9 & 21.2 & 198.4 \\
    Up-Down & ANN & 80.2 & 64.1 & 49.1 & 36.9 & 27.6 & 57.1 & 117.9 & 21.4 & 243.1 \\
    GCN-LSTM & ANN & 80.8 & 65.5 & 50.8 & 38.7 & 28.5 & 58.5 & 125.3 & 22.0 & 323.9 \\
    AoANet & ANN & 81.0 & 65.8 & 51.4 & 39.4 & 29.1 & 58.9 & 126.9 & 22.4 & 286.5 \\
    X-LAN & ANN & 80.8 & - & - & 39.5 & 29.5 & 59.2 & 132.0 & 23.4 & 210.7 \\ \hline
    X-Transformer & ANN & 81.9 & 66.9 & 52.4 & 40.3 & 29.6 & 59.5 & 131.1 & 23.4 & 368.4 \\
    $M^2$ Transformer & ANN & 81.6 & 66.4 & 51.8 & 39.7 & 29.4 & 59.2 & 129.3 & 22.6 & 280.2 \\
    RSTNet & ANN & 82.1 & 67.0 & 52.2 & 40.0 & 29.6 & 59.5 & 131.9 & 23.3 & 176.8 \\
    GET & ANN & 81.6 & 66.5 & 51.9 & 39.7 & 29.4 & 59.1 & 130.3 & 22.8 & 150.6 \\
    DLCT & ANN & 82.4 & 67.4 & 52.8 & 40.6 & 29.8 & 59.8 & 133.3 & 23.0 & 134.5 \\ \hline
    PureT & ANN & 82.8 & 68.1 & 53.6 & 41.4 & 30.1 & 60.4 & 136.0 & 24.2 & 387.9 \\
    
    \rowcolor{lightgray} SMM-Transformer-Large & SNN & 83.50 & 65.56 & 56.48 & 45.33 & 34.25 & 61.18 & 128.72 & 21.87 & \textbf{22.1} \\
    \bottomrule
    \end{tabular}}
    \label{tab:image_captioning_performance}
\end{table*}

We evaluate cross-modal retrieval on the COCO and Flickr30K datasets using the standard Karpathy split. SMM uses a dual-encoder retrieval architecture, enabling efficient image-text retrieval through dot-product similarity rather than exhaustive fusion-encoder scoring.
Tab.~\ref{retrieve} shows that SMM-Transformer-Base remains competitive among base-size models, while the Large variant further improves retrieval performance. We note that ALBEF-Base is stronger on some metrics, such as COCO R@1, because it uses a dense fusion encoder to rerank top-$k$ candidates.
In contrast, SMM focuses on a spike-compatible retrieval pathway with sparse token mixing and efficient dual-encoder inference.

\subsubsection{Image Caption}
As shown in Tab.~\ref{tab:image_captioning_performance}, SMM Transformer achieves competitive captioning quality while reducing estimated compute energy.
In particular, SMM-Transformer-Large attains BLEU-1/2/3/4 of 83.50/65.56/56.48/45.33, METEOR 34.25, ROUGE-L 61.18, CIDEr 128.72, and SPICE 21.87, with an estimated compute energy of 22.1\,mJ.
We note that CIDEr/SPICE remain below the best ANN baseline, suggesting that spike binarization and linearized token mixing may trade some semantic consensus for efficiency.
Fig.~\ref{fig:results} further provides qualitative evidence that SMM captures fine-grained visual cues and generates semantically faithful captions.

\subsubsection{Vision Tasks}

We evaluate SMM, employed as an image-only encoder, on two standard benchmarks: image classification (ImageNet) and semantic segmentation (ADE20K). As detailed in Table \ref{vision}, SMM exhibits competitive performance and notably surpasses the SNN-based PSSD-Transformer. The image resolution for ImageNet is 224×224, and for ADE20K, it is 512×512. We perform intermediate fine-tuning on ImageNet-21k for all four models.

\begin{table}[t]
\centering
\caption{Results on image classification and semantic segmentation.}
\label{vision}
\scalebox{0.85}{
\begin{tabular}{l|c|c|c}
\hline
\textbf{Models} & \textbf{Type} & \textbf{ImageNet (acc@1)} & \textbf{ADE20K (mIoU)} \\
\hline
ViT-Base & ANN & 83.6 & - \\
BEiT-Base & ANN & 85.2 & 52.8 \\
PSSD & SNN & 66.8 & 29.1 \\
SMM-Base & SNN  & 83.2 & 49.8 \\
\hline
\end{tabular}
}
\end{table}

\begin{table*}[t]
\centering
\caption{
Ablation studies of SMM Transformer and vision-language pre-training tasks. 
``ITC'' is short for image-text contrastive loss, ``ITM'' is image-text matching, and ``MLM'' is masked language modeling. The average of R@1, R@5 and R@10 is reported for Flickr30k. Results of NLVR2 are averaged over three runs.
}
\label{balbation}
\small
\begin{tabular}{lccc|ccc|cccc}
\toprule
& \multicolumn{3}{c}{\textbf{Pre-Training Tasks}} &
\multicolumn{3}{c}{\textbf{Transformer}} &
\multicolumn{2}{c}{\textbf{NLVR2}} & \multicolumn{2}{c}{\textbf{Flickr30k}} \\
& ITC & ITM & MLM & PLMP & SMSA & SMoE & dev & test-P & TR & IR \\
\midrule
\tblidx{1} & \cmark & \xmark  &\xmark &\cmark & \cmark &\cmark & 54.74 & 56.18 & 90.49 & 82.24  \\
\tblidx{2} & \cmark & \xmark & \cmark & \cmark& \cmark &\cmark & 72.53 & 70.56 & 91.60 & 84.29  \\
\tblidx{3} & \cmark & \cmark & \xmark & \cmark& \cmark &\cmark & 73.63 & 72.44 & 92.46 & 82.53  \\
\tblidx{4} & \cmark & \cmark & \cmark & \xmark &\cmark & \cmark& 76.74 & 76.89 & 90.72 & 84.27  \\
\tblidx{5} & \cmark & \cmark & \cmark & \cmark &\xmark & \cmark & 77.45 & 76.46 & 91.67 & 83.46  \\
\tblidx{6} & \cmark & \cmark & \cmark & \cmark& \cmark & \xmark& \ 77.55 &  76.89 &  93.04 &  85.53  \\
\tblidx{7} & \cmark & \cmark & \cmark & \cmark& \cmark & \cmark& \ \textbf{78.36} &  \textbf{77.53} &  \textbf{94.61} &  \textbf{86.22}  \\
\bottomrule
\end{tabular}

\label{tbl:ablation:mome_tasks}
\end{table*}

\begin{table*}[t]
\centering
\caption{PLMP performance under different neuron configurations and temporal parameter initializations. The results compare static LIF, single-branch learnable LIF, and multi-branch PLMP variants on image captioning.}
\label{tab:plmp_config}
\small
\setlength{\tabcolsep}{4.5pt}
\renewcommand{\arraystretch}{1.05}
\resizebox{\textwidth}{!}{
\begin{tabular}{lcccccc}
\toprule
Neuron configuration & BLEU-1 & BLEU-4 & METEOR & ROUGE-L & CIDEr & SPICE \\
\midrule
LIF, $\alpha=0.5$, $V_{th}=0.6$
& 78.46 & 34.32 & 27.15 & 54.94 & 125.26 & 19.87 \\
PLMP, $K=1$, $\alpha$ init. $=0.5$, $V_{th}=0.6$
& 78.89 & 36.00 & 27.30 & 55.98 & 125.69 & 20.77 \\
PLMP, $K=3$, shared $\alpha$ init., $V_{th}=0.3/0.6/0.9$
& 79.64 & 36.60 & 27.92 & 57.15 & 127.72 & 21.09 \\
PLMP, $K=3$, multi-stage $\alpha$ init., $V_{th}=0.3/0.6/0.9$
& 79.77 & 36.53 & 28.19 & 57.80 & 126.81 & 21.57 \\
\bottomrule
\end{tabular}
}
\end{table*}

\subsection{Ablation Study}\label{sec:ablation}

\paragraph{Pre-training tasks.}
We conduct ablation studies to evaluate the impact of different vision-language pre-training tasks, with the results summarized in Tab.~\ref{balbation}. Compared with the model trained only with image-text contrastive learning, adding masked language modeling and image-text matching consistently improves both NLVR2 and Flickr30K performance. These results show that SMM Transformer benefits from the same complementary supervision signals as ANN-based multimodal Transformers: contrastive learning improves global image-text alignment, masked language modeling strengthens textual grounding, and image-text matching with hard negatives improves fine-grained cross-modal discrimination.

\paragraph{Modality-aware fusion.}
We evaluate the contribution of SMoE in Tab.~\ref{balbation}. Removing SMoE from the full SMM Transformer leads to consistent performance drops on both NLVR2 and Flickr30K, indicating that modality-aware expert routing is beneficial for vision-language fusion. This result suggests that a single shared feed-forward transformation is insufficient to fully handle the heterogeneous representations produced by image tokens, text tokens, and joint image-text tokens.

SMoE addresses this issue by allocating separate feed-forward capacity to vision, language, and vision-language representations. The vision and language experts preserve modality-specific transformations, while the vision-language expert provides additional capacity for joint cross-modal representations after spike-driven token mixing. In this way, SMoE complements SMSA: SMSA provides a shared sparse interaction pathway, whereas SMoE performs modality-aware nonlinear transformation in the feed-forward path. We further isolate the role of the shared SMSA pathway in Appendix~\ref{app:shared_smsa}. Compared with separate SMSA parameters for image and text tokens in the early layers, shared SMSA achieves better NLVR2 and Flickr30K retrieval performance. 

\paragraph{PLMP and SMSA.}
As shown in Tab.~\ref{balbation}, experiments 4 and 5 validate the effectiveness of PLMP and SMSA, respectively. In experiment 4, we replace PLMP with standard LIF neurons and train the network with STBP. In experiment 5, we replace SMSA with the SSA baseline. Both replacements degrade performance, indicating that learnable multi-branch spiking dynamics and compensated spike-driven token mixing are important for the final model.

To further isolate the source of the PLMP gain, we report controlled neuron-configuration ablations in Tab.~\ref{tab:plmp_config}. The single-branch PLMP variant improves over the static LIF baseline while using the same branch count, showing that learnable leakage and threshold parameters already reduce the dependence on manually fixed LIF hyperparameters. The $K=3$ PLMP variants further improve most captioning metrics, indicating that parallel branches provide richer temporal responses than a single membrane trajectory. The comparison between shared and multi-stage leakage initializations shows that initialization affects the training trajectory and final metrics moderately, but both variants remain consistently stronger than static LIF because the leakage factors and thresholds are adapted during optimization.

We further compare SMSA with dense softmax self-attention and intermediate spike-driven variants in Tab.~\ref{tab:smsa_expressiveness}. Dense softmax self-attention provides the strongest token-pair interaction, while SSA introduces a larger distribution shift. SMSA without self-compensation reduces this gap, and the full SMSA further improves both task performance and representation similarity to the dense control.

\begin{table}[t]
\centering
\caption{Expressiveness analysis of SMSA. Cosine similarity and JS/KL shift are computed with respect to the dense softmax self-attention control.}
\label{tab:smsa_expressiveness}
\small
\setlength{\tabcolsep}{3.5pt}
\renewcommand{\arraystretch}{1.05}
\resizebox{\columnwidth}{!}{
\begin{tabular}{lcccc}
\toprule
Method & NLVR2 & F30K IR & Cos. & JS/KL \\
\midrule
Dense Softmax-SA & 77.80 & 86.40 & 1.000 & 0.000 \\
SSA baseline & 76.46 & 83.46 & 0.882 & 0.064 \\
SMSA w/o comp. & 76.97 & 85.11 & 0.921 & 0.041 \\
SMSA full & 77.53 & 86.22 & 0.947 & 0.026 \\
\bottomrule
\end{tabular}
}
\end{table}

\subsection{Energy Consumption}\label{sec:energy}

\begin{table*}[t]
\centering
\caption{Operator-level arithmetic energy of one vanilla self-attention block and one SMSA block under the MAC/AC model. $E_{\mathrm{MAC}}$ and $E_{\mathrm{AC}}$ denote the energy costs of multiply-accumulate and accumulate operations, respectively.}
\label{tab:attention_energy}
\small
\setlength{\tabcolsep}{8.5pt}
\renewcommand{\arraystretch}{1.2}
\begin{tabular}{lcc}
\toprule
Type & Vanilla Self-Attention & SMSA \\
\midrule
$Q,K,V$ projection 
& $E_{\mathrm{MAC}}\cdot 3ND^2$ 
& $E_{\mathrm{AC}}\cdot T\cdot R_{\mathrm{SA1}}\cdot 3ND^2$ \\

Attention mixing 
& $E_{\mathrm{MAC}}\cdot 2N^2D$ 
& $E_{\mathrm{AC}}\cdot T\cdot R_{\mathrm{SA2}}\cdot ND$ \\

Scale 
& $E_{\mathrm{MAC}}\cdot N^2$ 
& -- \\

Softmax 
& $E_{\mathrm{MAC}}\cdot N^2$ 
& -- \\

Output linear 
& $E_{\mathrm{MAC}}\cdot F_L$ 
& $E_{\mathrm{AC}}\cdot T\cdot R_L\cdot F_L$ \\

MLP layer 1 
& -- 
& $E_{\mathrm{AC}}\cdot T\cdot R_{\mathrm{M1}}\cdot F_{\mathrm{LM1}}$ \\

MLP layer 2 
& -- 
& $E_{\mathrm{AC}}\cdot T\cdot R_{\mathrm{M2}}\cdot F_{\mathrm{LM2}}$ \\
\bottomrule
\end{tabular}
\end{table*}

We evaluate efficiency from both module-level and whole-model perspectives. At the module level, we estimate the arithmetic energy of the attention block using the MAC/AC accounting model with 45nm CMOS constants, where $E_{\mathrm{MAC}}=4.6\,\mathrm{pJ}$ and $E_{\mathrm{AC}}=0.9\,\mathrm{pJ}$. For spike-driven operations, binary presynaptic activations convert multiply-accumulate operations into conditional accumulations, and inactive spikes skip the corresponding synaptic updates. We therefore scale the AC cost by the measured non-zero spike ratio of each operator path.

Tab.~\ref{tab:attention_energy} summarizes the operator-level accounting. Here, $N$ is the sequence length, $D$ is the hidden dimension, and $T$ is the number of SNN simulation steps. $F_L$ denotes the arithmetic count of the output linear projection, and $F_{\mathrm{LM1}}$ and $F_{\mathrm{LM2}}$ denote those of the two self-compensation MLP layers. $R_{\mathrm{SA1}}$, $R_{\mathrm{SA2}}$, $R_L$, $R_{\mathrm{M1}}$, and $R_{\mathrm{M2}}$ are the measured non-zero spike ratios of the corresponding operator paths. For the scale and softmax terms, we use one scalar arithmetic operation per attention logit as a proxy, without modeling hardware-specific exponential or division units.

\begin{table}[t]
\centering
\caption{Empirical non-zero spike ratios used in the MAC/AC accounting. Lower values indicate stronger event sparsity.}
\label{tab:sparsity_ratio}
\small
\begin{tabular}{lccccc}
\toprule
Split & $R_{\mathrm{SA1}}$ & $R_{\mathrm{SA2}}$ & $R_L$ & $R_{\mathrm{M1}}$ & $R_{\mathrm{M2}}$ \\
\midrule
Train & 0.132 & 0.081 & 0.157 & 0.172 & 0.138 \\
Eval & 0.107 & 0.061 & 0.132 & 0.149 & 0.117 \\
\bottomrule
\end{tabular}
\end{table}

As shown in Tab.~\ref{tab:sparsity_ratio}, SMSA maintains low activation rates across projection, mixing, output, and compensation paths. Using these measured ratios, SMSA reduces the estimated operator-level arithmetic energy of the attention block by up to 97\% compared with vanilla self-attention under the same hidden size, sequence length, and simulation-step setting. This estimate reflects attention-block arithmetic cost only and does not include memory access, scheduling overhead, or hardware-specific kernel effects.

\begin{table}[t]
\centering
\caption{Whole-model efficiency profile. Latency is measured under the same implementation and batch setting. Energy per pair is estimated from arithmetic operation counts and excludes memory access and hardware scheduling overhead.}
\label{tab:whole_model_efficiency}
\small
\begin{tabular}{lccc}
\toprule
Metric & Dense & SMM & Change \\
\midrule
Vision FLOPs (G) & 58.9 & 46.9 & -20.4\% \\
Language FLOPs (G) & 42.1 & 34.9 & -17.1\% \\
Whole-model FLOPs (G) & 170.4 & 127.6 & -25.1\% \\
Latency / batch (ms) & 92.4 & 79.8 & -13.6\% \\
Estimated energy / pair (mJ) & 356 & 279 & -21.6\% \\
\bottomrule
\end{tabular}
\end{table}

We further report whole-model FLOPs, latency, and arithmetic energy estimates in Tab.~\ref{tab:whole_model_efficiency}. The end-to-end gains are smaller than the attention-block gains because FFN/SMoE, embedding, projection, and task-head computations still occupy a substantial fraction of the total computation. Thus, the module-level estimate demonstrates the arithmetic advantage of SMSA, while the whole-model profile gives a more conservative view of SMM efficiency.
\section{Conclusion}
This work introduces an SNN-based framework for multimodal learning. The proposed PLMP neuron, together with P-STBP, stabilizes learnable multi-branch spiking dynamics. SMSA replaces dense pairwise softmax attention with attention-inspired channel-wise spike co-activation and self-compensation, providing a practical accuracy-efficiency trade-off. SMoE further introduces modality-aware expert routing for vision-language fusion. Experiments across visual and multimodal tasks show that SMM Transformer achieves competitive accuracy compared with ANN baselines. Our analysis shows large operator-level arithmetic savings for the attention module and more moderate but consistent whole-model efficiency gains. Future work will study memory-aware deployment and real-device neuromorphic measurements.

\section*{Acknowledgments}
This work was partly supported by Ningbo Key R\&D Program (2025Z047) , Ningbo Major Application Demonstration Program (2025Z199) and Ningbo Youth Science and Technology Innovation Leading Talent Project (2024QL044).

\section*{Impact Statement}
This paper presents work whose goal is to advance the field of machine learning. There are many potential societal consequences of our work, none of which we feel must be specifically highlighted here.
\bibliography{icml2026}

\begin{thebibliography}{47}
\providecommand{\natexlab}[1]{#1}
\providecommand{\url}[1]{\texttt{#1}}
\expandafter\ifx\csname urlstyle\endcsname\relax
  \providecommand{\doi}[1]{doi: #1}\else
  \providecommand{\doi}{doi: \begingroup \urlstyle{rm}\Url}\fi

\bibitem[Aafaq et~al.(2019)Aafaq, Mian, Liu, Gilani, and Shah]{sa2}
Aafaq, N., Mian, A., Liu, W., Gilani, S.~Z., and Shah, M.
\newblock Video description: A survey of methods, datasets, and evaluation metrics.
\newblock \emph{ACM Computing Surveys (CSUR)}, 52\penalty0 (6):\penalty0 1--37, 2019.

\bibitem[Anderson et~al.(2018)Anderson, He, Buehler, Teney, Johnson, Gould, and Zhang]{38}
Anderson, P., He, X., Buehler, C., Teney, D., Johnson, M., Gould, S., and Zhang, L.
\newblock Bottom-up and top-down attention for image captioning and visual question answering.
\newblock In \emph{Proceedings of the IEEE conference on computer vision and pattern recognition}, pp.\  6077--6086, 2018.

\bibitem[Bohte et~al.(2000)Bohte, Kok, and La~Poutr{\'e}]{spikeprop}
Bohte, S.~M., Kok, J.~N., and La~Poutr{\'e}, J.~A.
\newblock Spikeprop: backpropagation for networks of spiking neurons.
\newblock In \emph{ESANN}, volume~48, pp.\  419--424. Bruges, 2000.

\bibitem[Caporale \& Dan(2008)Caporale and Dan]{STDP}
Caporale, N. and Dan, Y.
\newblock Spike timing--dependent plasticity: a hebbian learning rule.
\newblock \emph{Annu. Rev. Neurosci.}, 31:\penalty0 25--46, 2008.

\bibitem[Carlini et~al.(2021)Carlini, Tramer, Wallace, Jagielski, Herbert-Voss, Lee, Roberts, Brown, Song, Erlingsson, et~al.]{LLM3}
Carlini, N., Tramer, F., Wallace, E., Jagielski, M., Herbert-Voss, A., Lee, K., Roberts, A., Brown, T., Song, D., Erlingsson, U., et~al.
\newblock Extracting training data from large language models.
\newblock In \emph{30th USENIX Security Symposium (USENIX Security 21)}, pp.\  2633--2650, 2021.

\bibitem[Chen et~al.(2024)Chen, Peng, Huang, and Tian]{rl}
Chen, D., Peng, P., Huang, T., and Tian, Y.
\newblock Deep reinforcement learning with spiking q-learning, 2024.

\bibitem[Dalal \& Triggs(2005)Dalal and Triggs]{36}
Dalal, N. and Triggs, B.
\newblock Histograms of oriented gradients for human detection.
\newblock In \emph{2005 IEEE computer society conference on computer vision and pattern recognition (CVPR'05)}, volume~1, pp.\  886--893. Ieee, 2005.

\bibitem[Devlin et~al.(2019)Devlin, Chang, Lee, and Toutanova]{bert}
Devlin, J., Chang, M.-W., Lee, K., and Toutanova, K.
\newblock Bert: Pre-training of deep bidirectional transformers for language understanding.
\newblock In \emph{Proceedings of the 2019 conference of the North American chapter of the association for computational linguistics: human language technologies, volume 1 (long and short papers)}, pp.\  4171--4186, 2019.

\bibitem[Dosovitskiy(2020)]{vit}
Dosovitskiy, A.
\newblock An image is worth 16x16 words: Transformers for image recognition at scale.
\newblock \emph{arXiv preprint arXiv:2010.11929}, 2020.

\bibitem[Dubey et~al.(2024)Dubey, Jauhri, Pandey, Kadian, Al-Dahle, Letman, Mathur, Schelten, Yang, Fan, et~al.]{grattafiori2024llama3herdmodels}
Dubey, A., Jauhri, A., Pandey, A., Kadian, A., Al-Dahle, A., Letman, A., Mathur, A., Schelten, A., Yang, A., Fan, A., et~al.
\newblock The llama 3 herd of models.
\newblock \emph{arXiv e-prints}, pp.\  arXiv--2407, 2024.

\bibitem[Fang et~al.(2021)Fang, Yu, Chen, Huang, Masquelier, and Tian]{SEW-Resnet}
Fang, W., Yu, Z., Chen, Y., Huang, T., Masquelier, T., and Tian, Y.
\newblock Deep residual learning in spiking neural networks.
\newblock \emph{Advances in Neural Information Processing Systems}, 34:\penalty0 21056--21069, 2021.

\bibitem[Furber(2016)]{nnc2}
Furber, S.
\newblock Large-scale neuromorphic computing systems.
\newblock \emph{Journal of neural engineering}, 13\penalty0 (5):\penalty0 051001, 2016.

\bibitem[Goodfellow et~al.(2014)Goodfellow, Pouget-Abadie, Mirza, Xu, Warde-Farley, Ozair, Courville, and Bengio]{gan}
Goodfellow, I.~J., Pouget-Abadie, J., Mirza, M., Xu, B., Warde-Farley, D., Ozair, S., Courville, A., and Bengio, Y.
\newblock Generative adversarial networks, 2014.

\bibitem[Hebb(2005)]{hebb}
Hebb, D.~O.
\newblock \emph{The organization of behavior: A neuropsychological theory}.
\newblock Psychology press, 2005.

\bibitem[Herdade et~al.(2019)Herdade, Kappeler, Boakye, and Soares]{39}
Herdade, S., Kappeler, A., Boakye, K., and Soares, J.
\newblock Image captioning: Transforming objects into words.
\newblock \emph{Advances in neural information processing systems}, 32, 2019.

\bibitem[Jin et~al.(2018)Jin, Zhang, and Li]{DSNN3}
Jin, Y., Zhang, W., and Li, P.
\newblock Hybrid macro/micro level backpropagation for training deep spiking neural networks.
\newblock \emph{Advances in neural information processing systems}, 31, 2018.

\bibitem[Kaiser et~al.(2020)Kaiser, Mostafa, and Neftci]{DSNN4}
Kaiser, J., Mostafa, H., and Neftci, E.
\newblock Synaptic plasticity dynamics for deep continuous local learning (decolle).
\newblock \emph{Frontiers in Neuroscience}, 14:\penalty0 515306, 2020.

\bibitem[Kasneci et~al.(2023)Kasneci, Se{\ss}ler, K{\"u}chemann, Bannert, Dementieva, Fischer, Gasser, Groh, G{\"u}nnemann, H{\"u}llermeier, et~al.]{LLM1}
Kasneci, E., Se{\ss}ler, K., K{\"u}chemann, S., Bannert, M., Dementieva, D., Fischer, F., Gasser, U., Groh, G., G{\"u}nnemann, S., H{\"u}llermeier, E., et~al.
\newblock Chatgpt for good? on opportunities and challenges of large language models for education.
\newblock \emph{Learning and individual differences}, 103:\penalty0 102274, 2023.

\bibitem[Koley et~al.(2024)Koley, Bhunia, Sain, Chowdhury, Xiang, and Song]{koley2024text}
Koley, S., Bhunia, A.~K., Sain, A., Chowdhury, P.~N., Xiang, T., and Song, Y.-Z.
\newblock Text-to-image diffusion models are great sketch-photo matchmakers.
\newblock In \emph{Proceedings of the IEEE/CVF Conference on Computer Vision and Pattern Recognition}, pp.\  16826--16837, 2024.

\bibitem[Lei et~al.(2018)Lei, Yu, Bansal, and Berg]{sa5}
Lei, J., Yu, L., Bansal, M., and Berg, T.~L.
\newblock Tvqa: Localized, compositional video question answering.
\newblock \emph{arXiv preprint arXiv:1809.01696}, 2018.

\bibitem[Liu et~al.(2024)Liu, Li, Wu, Xu, and Yang]{liu2024multi}
Liu, W., Li, J., Wu, Z., Xu, J., and Yang, B.
\newblock Multi-head hashing with orthogonal decomposition for cross-modal retrieval.
\newblock In \emph{International Conference on Multimedia Modeling}, pp.\  170--183. Springer, 2024.

\bibitem[Lowe(1999)]{35}
Lowe, D.~G.
\newblock Object recognition from local scale-invariant features.
\newblock In \emph{Proceedings of the seventh IEEE international conference on computer vision}, volume~2, pp.\  1150--1157. Ieee, 1999.

\bibitem[Markovi{\'c} et~al.(2020)Markovi{\'c}, Mizrahi, Querlioz, and Grollier]{nnc1}
Markovi{\'c}, D., Mizrahi, A., Querlioz, D., and Grollier, J.
\newblock Physics for neuromorphic computing.
\newblock \emph{Nature Reviews Physics}, 2\penalty0 (9):\penalty0 499--510, 2020.

\bibitem[Pan et~al.(2020)Pan, Yao, Li, and Mei]{40}
Pan, Y., Yao, T., Li, Y., and Mei, T.
\newblock X-linear attention networks for image captioning.
\newblock In \emph{Proceedings of the IEEE/CVF conference on computer vision and pattern recognition}, pp.\  10971--10980, 2020.

\bibitem[Rathi et~al.(2020)Rathi, Srinivasan, Panda, and Roy]{DSNN5}
Rathi, N., Srinivasan, G., Panda, P., and Roy, K.
\newblock Enabling deep spiking neural networks with hybrid conversion and spike timing dependent backpropagation.
\newblock \emph{arXiv preprint arXiv:2005.01807}, 2020.

\bibitem[Ren et~al.(2015)Ren, Kiros, and Zemel]{sa3}
Ren, M., Kiros, R., and Zemel, R.
\newblock Exploring models and data for image question answering.
\newblock \emph{Advances in neural information processing systems}, 28, 2015.

\bibitem[Roy et~al.(2019)Roy, Jaiswal, and Panda]{2watts}
Roy, K., Jaiswal, A., and Panda, P.
\newblock Towards spike-based machine intelligence with neuromorphic computing.
\newblock \emph{Nature}, 575\penalty0 (7784):\penalty0 607--617, 2019.

\bibitem[Sengupta et~al.(2019)Sengupta, Ye, Wang, and Roy]{ANN2SNN1}
Sengupta, A., Ye, Y., Wang, R., and Roy, K.
\newblock Going deeper in spiking neural networks: Vgg and residual architectures.
\newblock \emph{Frontiers in neuroscience}, 13:\penalty0 425055, 2019.

\bibitem[Suo et~al.(2024)Suo, Ma, Zhu, and Yang]{suo2024knowledge}
Suo, Y., Ma, F., Zhu, L., and Yang, Y.
\newblock Knowledge-enhanced dual-stream zero-shot composed image retrieval.
\newblock In \emph{Proceedings of the IEEE/CVF Conference on Computer Vision and Pattern Recognition}, pp.\  26951--26962, 2024.

\bibitem[Tavanaei et~al.(2019)Tavanaei, Ghodrati, Kheradpisheh, Masquelier, and Maida]{LifNode}
Tavanaei, A., Ghodrati, M., Kheradpisheh, S.~R., Masquelier, T., and Maida, A.
\newblock Deep learning in spiking neural networks.
\newblock \emph{Neural networks}, 111:\penalty0 47--63, 2019.

\bibitem[van De~Burgt et~al.(2018)van De~Burgt, Melianas, Keene, Malliaras, and Salleo]{nnc3}
van De~Burgt, Y., Melianas, A., Keene, S.~T., Malliaras, G., and Salleo, A.
\newblock Organic electronics for neuromorphic computing.
\newblock \emph{Nature Electronics}, 1\penalty0 (7):\penalty0 386--397, 2018.

\bibitem[Vaswani et~al.(2017)Vaswani, Shazeer, Parmar, Uszkoreit, Jones, Gomez, Kaiser, and Polosukhin]{Transformer}
Vaswani, A., Shazeer, N., Parmar, N., Uszkoreit, J., Jones, L., Gomez, A.~N., Kaiser, {\L}., and Polosukhin, I.
\newblock Attention is all you need.
\newblock \emph{Advances in neural information processing systems}, 30, 2017.

\bibitem[Wang et~al.(2025)Wang, Li, Zhu, Li, Zhang, and Shen]{wang2025cross}
Wang, T., Li, F., Zhu, L., Li, J., Zhang, Z., and Shen, H.~T.
\newblock Cross-modal retrieval: a systematic review of methods and future directions.
\newblock \emph{Proceedings of the IEEE}, 2025.

\bibitem[Wang et~al.(2022)Wang, Xu, and Sun]{41}
Wang, Y., Xu, J., and Sun, Y.
\newblock End-to-end transformer based model for image captioning.
\newblock In \emph{Proceedings of the AAAI Conference on Artificial Intelligence}, volume~36, pp.\  2585--2594, 2022.

\bibitem[Wei et~al.(2022)Wei, Tay, Bommasani, Raffel, Zoph, Borgeaud, Yogatama, Bosma, Zhou, Metzler, et~al.]{LLM2}
Wei, J., Tay, Y., Bommasani, R., Raffel, C., Zoph, B., Borgeaud, S., Yogatama, D., Bosma, M., Zhou, D., Metzler, D., et~al.
\newblock Emergent abilities of large language models.
\newblock \emph{arXiv preprint arXiv:2206.07682}, 2022.

\bibitem[Wu et~al.(2018)Wu, Deng, Li, and Shi]{STBP}
Wu, Y., Deng, L., Li, G., and Shi, L.
\newblock Spatio-temporal backpropagation for training high-performance spiking neural networks.
\newblock \emph{Frontiers in neuroscience}, 12:\penalty0 323875, 2018.

\bibitem[Xiao et~al.(2022)Xiao, Zhou, Chua, and Yan]{sa6}
Xiao, J., Zhou, P., Chua, T.-S., and Yan, S.
\newblock Video graph transformer for video question answering.
\newblock In \emph{European Conference on Computer Vision}, pp.\  39--58. Springer, 2022.

\bibitem[Xie et~al.(2024)Xie, Wang, Xie, Tan, Li, Li, Peng, Tang, and Fang]{xie2024image}
Xie, Y., Wang, Y., Xie, Y., Tan, X., Li, J., Li, X., Peng, W., Tang, M., and Fang, M.
\newblock Image-text retrieval with main semantics consistency.
\newblock In \emph{Proceedings of the 33rd ACM International Conference on Information and Knowledge Management}, pp.\  2629--2638, 2024.

\bibitem[Xu et~al.(2016)Xu, Mei, Yao, and Rui]{sa1}
Xu, J., Mei, T., Yao, T., and Rui, Y.
\newblock Msr-vtt: A large video description dataset for bridging video and language.
\newblock In \emph{Proceedings of the IEEE conference on computer vision and pattern recognition}, pp.\  5288--5296, 2016.

\bibitem[Xu et~al.(2015)Xu, Ba, Kiros, Cho, Courville, Salakhudinov, Zemel, and Bengio]{37}
Xu, K., Ba, J., Kiros, R., Cho, K., Courville, A., Salakhudinov, R., Zemel, R., and Bengio, Y.
\newblock Show, attend and tell: Neural image caption generation with visual attention.
\newblock In \emph{International conference on machine learning}, pp.\  2048--2057. PMLR, 2015.

\bibitem[Yang et~al.(2017)Yang, Tang, Yang, and Li]{44}
Yang, L., Tang, K., Yang, J., and Li, L.-J.
\newblock Dense captioning with joint inference and visual context.
\newblock In \emph{Proceedings of the IEEE conference on computer vision and pattern recognition}, pp.\  2193--2202, 2017.

\bibitem[Yang et~al.(2016)Yang, He, Gao, Deng, and Smola]{sa4}
Yang, Z., He, X., Gao, J., Deng, L., and Smola, A.
\newblock Stacked attention networks for image question answering.
\newblock In \emph{Proceedings of the IEEE conference on computer vision and pattern recognition}, pp.\  21--29, 2016.

\bibitem[Yao et~al.(2024)Yao, Hu, Zhou, Yuan, Tian, Xu, and Li]{Spike-driven}
Yao, M., Hu, J., Zhou, Z., Yuan, L., Tian, Y., Xu, B., and Li, G.
\newblock Spike-driven transformer.
\newblock \emph{Advances in Neural Information Processing Systems}, 36, 2024.

\bibitem[Yu et~al.(2021)Yu, Ma, Song, Zhang, Dang, and Tan]{ANN2SNN3}
Yu, Q., Ma, C., Song, S., Zhang, G., Dang, J., and Tan, K.~C.
\newblock Constructing accurate and efficient deep spiking neural networks with double-threshold and augmented schemes.
\newblock \emph{IEEE Transactions on Neural Networks and Learning Systems}, 33\penalty0 (4):\penalty0 1714--1726, 2021.

\bibitem[Zeng et~al.(2024)Zeng, Xie, Zhang, Chen, Chen, and Wang]{zeng2024meacap}
Zeng, Z., Xie, Y., Zhang, H., Chen, C., Chen, B., and Wang, Z.
\newblock Meacap: Memory-augmented zero-shot image captioning.
\newblock In \emph{Proceedings of the IEEE/CVF conference on computer vision and pattern recognition}, pp.\  14100--14110, 2024.

\bibitem[Zhang \& Li(2019)Zhang and Li]{DSNN7}
Zhang, W. and Li, P.
\newblock Spike-train level backpropagation for training deep recurrent spiking neural networks.
\newblock \emph{Advances in neural information processing systems}, 32, 2019.

\bibitem[Zhou et~al.(2022)Zhou, Zhu, He, Wang, Yan, Tian, and Yuan]{spikformer}
Zhou, Z., Zhu, Y., He, C., Wang, Y., Yan, S., Tian, Y., and Yuan, L.
\newblock Spikformer: When spiking neural network meets transformer.
\newblock \emph{arXiv preprint arXiv:2209.15425}, 2022.

\end{thebibliography}
\bibliographystyle{icml2026}

\newpage
\appendix
\onecolumn

\section{Derivations for PLMP and P-STBP}\label{app:plmp_pstbp}

This section provides the error-propagation details omitted from the main text. The forward dynamics of PLMP, including branch membrane updates, branch spikes, internal spike-count aggregation, and binary read-out spikes, have already been defined in Sec.~2.

P-STBP uses surrogate derivatives for the non-differentiable branch spike and read-out spike functions:
\begin{equation}
    \frac{\partial o_{i,k}^{t,n}}
    {\partial u_{i,k}^{t,n}}
    \approx
    \phi_{\mathrm{br}}
    \left(
    u_{i,k}^{t,n}
    -
    V_{th,k}
    \right),
    \quad
    \frac{\partial \hat{o}_{i}^{t,n}}
    {\partial c_i^{t,n}}
    \approx
    \phi_{\mathrm{ro}}
    \left(
    c_i^{t,n}
    -
    \theta_{\mathrm{ro}}
    \right).
\end{equation}
Here, $\phi_{\mathrm{br}}(\cdot)$ is used for branch spikes, while $\phi_{\mathrm{ro}}(\cdot)$ is used for the fixed-threshold PLMP read-out gate. The task-specific loss provides the upstream gradient at the output layer. For hidden PLMP layers, the read-out spike receives gradients from the next layer:
\begin{align}
    \frac{\partial L}{\partial \hat{o}_{i}^{t,n}}
    &=
    \sum_{j\in \mathcal{N}_{n+1}}
    \sum_{k=1}^{K}
    \frac{\partial L}{\partial u_{j,k}^{t,n+1}}
    w_{ji}^{n+1},
    \notag \\
    \frac{\partial L}{\partial o_{i,k}^{t,n}}
    &=
    \frac{\partial L}{\partial \hat{o}_{i}^{t,n}}
    \phi_{\mathrm{ro}}
    \left(
    c_i^{t,n}
    -
    \theta_{\mathrm{ro}}
    \right)
    -
    \frac{\partial L}{\partial u_{i,k}^{t+1,n}}
    \alpha_k
    u_{i,k}^{t,n},
    \notag \\
    \frac{\partial L}{\partial u_{i,k}^{t,n}}
    &=
    \frac{\partial L}{\partial o_{i,k}^{t,n}}
    \phi_{\mathrm{br}}
    \left(
    u_{i,k}^{t,n}
    -
    V_{th,k}
    \right)
    +
    \frac{\partial L}{\partial u_{i,k}^{t+1,n}}
    \alpha_k
    \left(
    1-o_{i,k}^{t,n}
    \right).
\end{align}
The first line corresponds to spatial error propagation across layers, where $\mathcal{N}_{n+1}$ denotes the set of postsynaptic neurons in the next layer. The second line contains two paths for the branch spike: the read-out path at the current time step and the reset path that affects the next membrane state. The third line combines the current branch-spike surrogate gradient and the temporal recurrent gradient from the next simulation step.

Since the read-out spike $\hat{o}_{i}^{t,n}$ is generated from the internal spike count $c_i^{t,n}$, gradients must first pass through the read-out gate and then be distributed to the $K$ branch spikes. Meanwhile, each branch spike also affects the next membrane state through the reset term in the PLMP membrane update. Therefore, P-STBP propagates errors not only across layers and time steps, but also across the internal branch aggregation structure of PLMP.

If the learnable leakage factor is represented by an unconstrained parameter $m_k$, we use $\alpha_k=\operatorname{sigmoid}(m_k)$ and apply the chain rule:
\begin{equation}
    \frac{\partial L}{\partial m_k}
    =
    \frac{\partial L}{\partial \alpha_k}
    \alpha_k
    \left(
    1-\alpha_k
    \right).
\end{equation}
This parameterization keeps the leakage factor in a stable range during training. 
These equations clarify why P-STBP differs from standard STBP. Standard STBP propagates through a single LIF membrane trajectory, whereas P-STBP must propagate through the temporal membrane recursion, the branch spike surrogate, the internal spike-count aggregation, and the binary read-out gate. This allows PLMP to learn branch-specific leakage factors and thresholds while propagating only binary read-out spikes to subsequent layers.

\section{Additional Experiments and Statistics}\label{app:extra_exp}

This section provides additional sensitivity results on the number of PLMP branches and further analyzes the shared SMSA design.

\begin{table}[h]
\centering
\caption{Sensitivity of PLMP to the number of parallel LIF branches. Cost is normalized by the static LIF baseline.}
\label{tab:plmp_k_sensitivity}
\small
\begin{tabular}{lccc}
\toprule
Neuron configuration & BLEU-4 & CIDEr & Cost \\
\midrule
LIF & 34.32 & 125.26 & 1.00$\times$ \\
Learnable-LIF & 35.84 & 126.18 & 1.06$\times$ \\
PLMP, $K=2$ & 36.34 & 126.94 & 1.12$\times$ \\
PLMP, $K=3$ & 36.60 & 127.72 & 1.17$\times$ \\
PLMP, $K=4$ & 36.56 & 127.63 & 1.24$\times$ \\
\bottomrule
\end{tabular}
\end{table}

\begin{figure}[t]
    \centering
    \includegraphics[width=0.8\linewidth]{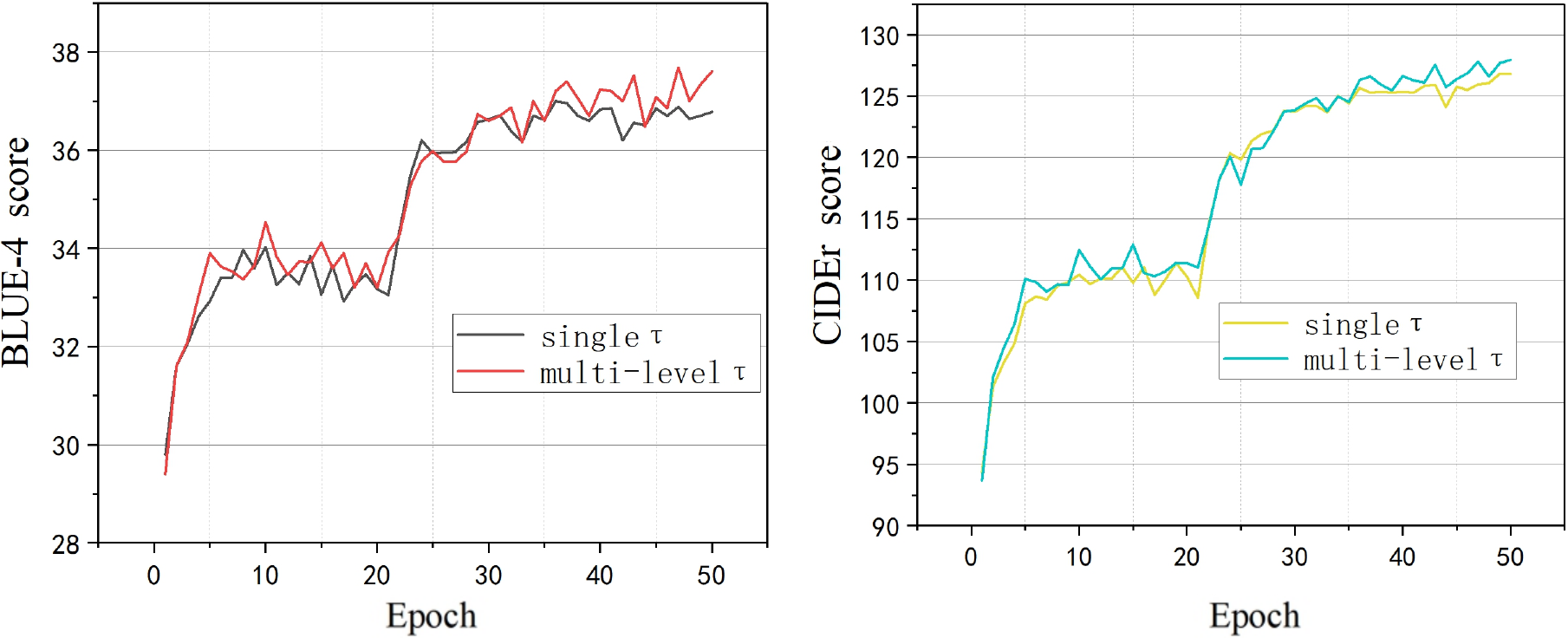}
    \caption{Training curves for PLMP variants with different temporal parameter initializations.}
    \label{fig:zhexian}
\end{figure}

\begin{table}[t]
\centering
\caption{Controlled ablation of shared SMSA in the early multimodal layers. Separate SMSA uses different spike-driven token-mixing parameters for image patches and text tokens in the first $L-F$ layers, while shared SMSA uses a common SMSA pathway before modality-aware SMoE transformations.}
\label{tab:shared_smsa}
\small
\begin{tabular}{lcccc}
\toprule
Setting & NLVR2 dev & NLVR2 test-P & Flickr30K TR & Flickr30K IR \\
\midrule
Separate SMSA & 77.53 & 76.58 & 93.21 & 85.98 \\
Shared SMSA + SMoE & 79.17 & 79.42 & 94.57 & 86.33 \\
\bottomrule
\end{tabular}
\end{table}

Tab.~\ref{tab:plmp_k_sensitivity} studies the sensitivity of PLMP to the number of parallel branches. Increasing $K$ from 1 to 3 consistently improves BLEU-4 and CIDEr, indicating that multiple temporal branches are useful for representing diverse firing dynamics. However, increasing $K$ from 3 to 4 brings negligible performance improvement while increasing computational cost. Therefore, we use $K=3$ by default in the main experiments, as it provides the best accuracy-cost trade-off.

Fig.~\ref{fig:zhexian} compares PLMP variants with different temporal parameter initializations. The curves show that learnable temporal parameters reduce sensitivity to the initial leakage configuration. Although different initializations lead to slightly different early training trajectories, the final performance remains close after optimization. This supports the motivation of PLMP: branch-specific learnable temporal parameters allow the neuron to adapt its firing dynamics during training instead of relying on a manually fixed membrane constant.

\subsection{Shared SMSA versus Separate SMSA}\label{app:shared_smsa}

Tab.~\ref{tab:shared_smsa} studies whether image and text tokens should use separate SMSA parameters or a shared SMSA pathway in the early layers. The separate variant encodes image patches and text tokens with different SMSA parameters before fusion, whereas the shared variant exposes both modalities to a common spike-driven token-mixing process. The shared design improves both NLVR2 and Flickr30K retrieval, suggesting that early common SMSA helps visual and textual tokens form a more aligned representation space before the modality-aware SMoE feed-forward transformations. This ablation complements Tab.~\ref{tab:smsa_expressiveness}. The expressiveness analysis studies how self-compensation reduces the representation gap between SMSA and dense softmax self-attention, while the shared-SMSA ablation studies whether a common early token-mixing pathway is beneficial for cross-modal alignment.

\end{document}